\documentclass[9pt,twocolumn,twoside,lineno]{pnas-new}
\usepackage{amsmath}
\usepackage{amsfonts}
\usepackage{amssymb}
\usepackage{float}
\usepackage{listings}
\usepackage{multicol}
\usepackage{varioref}
\usepackage{enumitem}   
\usepackage{graphicx,wrapfig,lipsum}
\usepackage{todonotes}

\usepackage[xindy,toc]{glossaries}
\usepackage{color}
\usepackage{caption}
\usepackage{verbatim}
\usepackage{subcaption}
\usepackage{bussproofs}
\usepackage{stmaryrd}
\usepackage{faktor}
\usepackage{enumitem}
\usepackage{amssymb}
\usepackage{epigraph}

\templatetype{pnasresearcharticle} % Choose template 
\title{Classifying Organisms and Artefacts By Their Shapes}
\author[a]{Arianna Salili-James}
\author[b]{Anne Mackay}
\author[c]{Emilio Rodriguez-Alvarez}
\author[d]{Diana Rodriguez-Perez}
\author[d]{Thomas Mannack}
\author[e]{Timothy A. Rawlings}
\author[f]{A. Richard Palmer}
\author[g]{Jonathan Todd}
\author[h]{Terhi E. Riutta}
\author[i,j]{Cate Macinnis-Ng}
\author[h]{Zhitong Han}
\author[h]{Megan Davies}
\author[h]{Zinnia Thorpe}
\author[l,j]{Stephen Marsland}
\author[h]{Armand M. Leroi}

\affil[a]{Department of Mathematics, Brunel University London, Uxbridge, UB8 3PH, UK}
\affil[b]{School of Humanities, University of Auckland, Auckland 1010, New Zealand}
\affil[c]{School of Anthropology, University of Arizona, Tucson, AZ 85721-0030, USA}
\affil[d]{Classical Art Research Centre, Ioannou Centre for Classical and Byzantine Studies, University of Oxford,  Oxford, OX1 3LU, UK}
\affil[e]{School of Science \& Technology, Cape Breton University, Sydney, Nova Scotia, B1P 6L2, Canada}
\affil[f]{Department of Biological Sciences
University of Alberta, Edmonton, Alberta, T6G 2E9,  Canada}
\affil[g]{Department of Earth Sciences, Natural History Museum, London, SW7 5BD, UK}
\affil[h]{Department of Life Sciences, Imperial College London, London, SW7 2AZ, UK}
\affil[i]{School of Biological Sciences, University of Auckland, Auckland 1010, New Zealand}
\affil[j]{Te P\=unaha Matatini, New Zealand}
\affil[l]{School of Mathematics and Statistics, Victoria University of Wellington, Wellington 6012, New Zealand}

\leadauthor{Leroi \& Marsland} 

\significancestatement{Classifying objects by their shapes is useful in many scientific fields. However, the mathematical space of shapes is non-linear and complicated. We ask whether or not this matters for three datasets of interest to social and natural scientists: classifying the outlines of vases, leaves and shells. We compare a traditional method of shape analysis based on linear analysis of points that represent the shape with three diffeomorphic shape methods based explicitly on the curves that deal with true shape spaces. Our experiments suggest that diffeomorphic shape analysis methods should be part of the toolbox of the scientist, and that they even outperform human experts on this task. Finally, we show that the intermediary shapes can also suggest potential evolutionary pathways.
}

\authorcontributions{SM and AML designed the study, with input from AS-J; MD, ZH and ZT collected data; AM, TM, DR-P, and TER vetted data; AM, ER-A, DR-P, TAR, ARP, JT, TER, and CM-N were expert classifiers; AS-J and SM prepared the data and implemented shape analysis methods; AS-J, SM and AML analysed data; SM and AML wrote the paper with contributions from AS-J; all authors edited it.}
\authordeclaration{The authors declare no competing interests}
\equalauthors{\textsuperscript{1}SM and AML contributed equally to this work.}
\correspondingauthor{\textsuperscript{2}To whom correspondence should be addressed. E-mail: a.leroi@imperial.ac.uk or stephen.marsland@vuw.ac.nz}


\begin{abstract}

We often wish to classify objects by their shapes. Indeed, the study of shapes is an important part of many scientific fields such as evolutionary biology, structural biology, image processing, and archaeology. The most widely-used method of shape analysis, Geometric Morphometrics, assumes that that the mathematical space in which shapes are represented is linear.  However, it has long been known that shape space is, in fact, rather more complicated, and certainly non-linear.  Diffeomorphic methods that take this non-linearity into account, and so give more accurate estimates of the distances among shapes, exist but have rarely been applied to real-world problems. Using a machine classifier, we tested the ability of several of these methods to describe and classify the shapes of a variety of organic and man-made objects. We find that one method, the Square-Root Velocity Function (SRVF), is superior to all others, including a standard Geometric Morphometric method (eigenshapes). We also show that computational shape classifiers outperform human experts, and that the SRVF shortest-path between shapes can be used to estimate the shapes of intermediate steps in evolutionary series. Diffeomorphic shape analysis methods, we conclude, now provide practical and effective solutions to many shape description and classification problems in the natural and human sciences.
\end{abstract}

\dates{This manuscript was compiled on \today}
\doi{\url{www.pnas.org/cgi/doi/10.1073/pnas.XXXXXXXXXX}}

\begin{document}

\maketitle
\thispagestyle{firststyle}
\ifthenelse{\boolean{shortarticle}}{\ifthenelse{\boolean{singlecolumn}}{\abscontentformatted}{\abscontent}}{}

%, that is, assemble them into similar groups,
\dropcap{G}iven a set of images of objects, we may wish to classify them by their \emph{shapes}, by which we mean their forms stripped of any differences in size, orientation, position in space, or surface patterns. Humans intuitively understand shape in this sense: we identify objects that have the same shape even when they differ in size or are oriented at different angles relative to us \cite{Mumford1991,Small,Dryden, Marr,Sutherland}, and other animals seem to have similar abilities \cite{Mumford1989,Tang2004}. The analysis and classification of shapes has applications in fields as varied as biology, medicine, archaeology, image analysis, and architecture \cite{Bookstein,Zelditch12,Dryden}; many algorithmic methods that allow us to do so objectively have, accordingly, been proposed. These methods differ in how they describe shapes and estimate the distances among them.  A shape outline can be represented in two fundamentally different ways: either as a set of points or else as a curve. If represented as sets of corresponding points then the distance between two shapes is typically computed as the sum of the distances between the points; if represented as curves then as the amount the amount of bending and stretching required to transform one into the other \cite{Ramsay2002,Srivastava16}. Insofar that these two approaches differ not only in how they represent shapes, but in the  distance metrics that they use, they also make different assumptions about the geometry of shape space. 

\section*{The strange geometry of shape space}

\begin{figure}
\centering
\includegraphics[width=.8\linewidth]{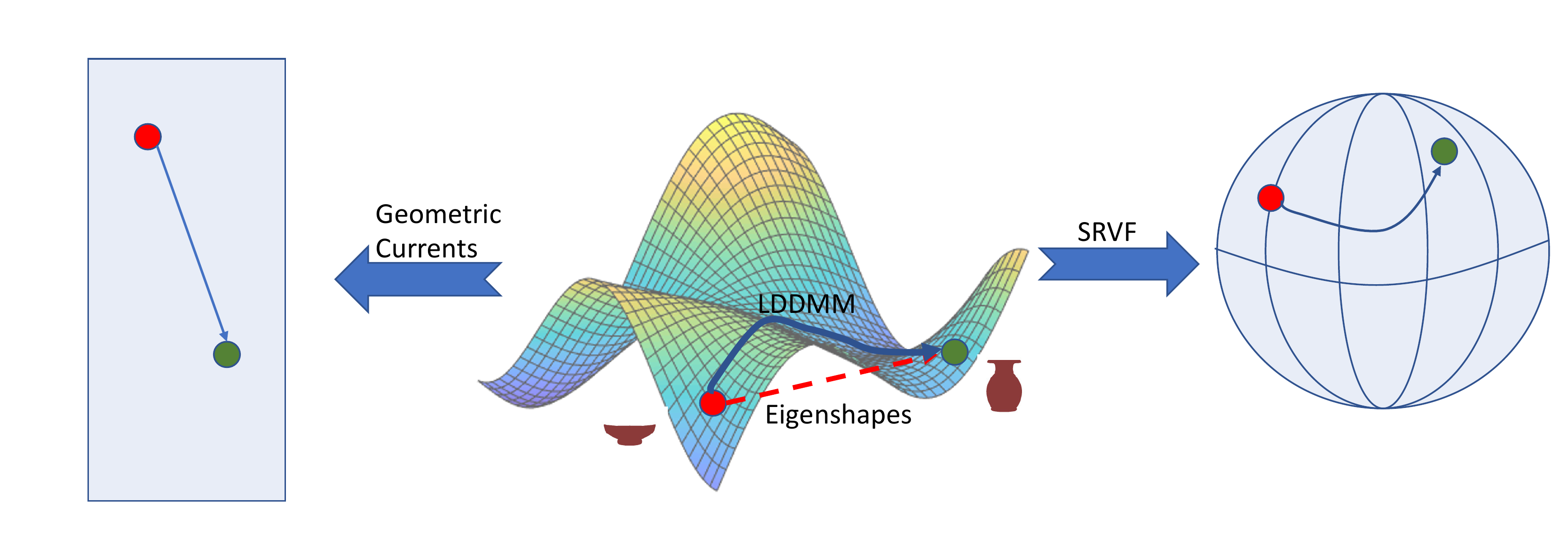}
\caption{An overview of shape analysis methods. Shapes, such as the two vases shown, are described by points in a manifold of possible shapes (centre). Eigenshape analysis ignores the complex geometry of shape space and assumes that distances are linear. Diffeomorphic methods, by contrast, either work on the original shape space (e.g., Large Deformation Diffeomorphic Metric Mapping (LDDMM)) or else transform the original shapes into a simpler space --- a sphere in the case of Square-Root Velocity Function (SRVF) or a vector space in the case of Geometric Currents (GC).}
\label{fig:methods}
\end{figure}

Although a planar shape can be represented as a curve drawn in two dimensional space, the space of $N$ point positions representing that curve has many more dimensions: for each of the $N$ points we have $x$ and $y$ coordinates, so need $2N$ numbers to describe each shape. Point configurations that differ only in position, scale, reflection, rotation, or some combination of these, describe the same shape, which means that the set of different \emph{shapes} is not the whole of $\mathbb{R}^{2N}$, but a subspace of it. And that, in turn, means that shapes live not in ordinary Euclidean space but in a non-linear space embedded within it \cite{Kendall77,Kendall1984}.  For the basic case of triangles -- three points in $\mathbb{R}^6$ -- the shape space of allowable shapes is a hemisphere \cite{Small,Klingenberg}. The points describing more complex shapes exist in unknown, but certainly much more complex, shape spaces (Figure \ref{fig:methods}).  And curves, which are continuous, exist in infinite dimensional shape spaces \cite{Younes}. 

Geometric Morphometric methods treat outline shapes as sets of points. As such they are \emph{linear} methods that operate in Euclidean space \cite{Bookstein,Zelditch12,Macleod2018}. Known variously as eigenshape analysis \cite{Lohmann83,MacLeod_EA,Macleod1999} or Statistical Shape Models \cite{ASM}, they begin by extracting shapes from forms by standardizing position, scale, and rotation using a procedure called Procrustes alignment \cite{Gower}. The dimensionality of the space of coordinate points is then reduced by a method such as Principal Components Analysis, and the distances between shapes computed among the vectors of derived variables. 

The strange geometry of shape space means, however, that distances between sets of point coordinates may not be very good estimators of the true distances between shapes. An alternative is to consider the shape as a piece of elastic and ask how much `energy' --- stretching and bending --- is required to transform one shape into another. Under certain restrictions, one curve can be continuously deformed into another using a smooth, invertible, function called a \emph{diffeomorphism}. Large Deformation Diffeomorphic Metric Mapping (LDDMM) algorithms transform shape curves into each other and estimate a distance directly in the space of diffeomorphisms, which is an infinite-dimensional manifold that can be equipped with a Riemannian metric \cite{Beg05,Younes} (Figure \ref{fig:methods}). As with Geometric Morphometrics, they deal with form not shape, and so require that Procrustes alignment be applied first. 

It can be beneficial to simplify the shape space by transforming the shapes before analysis. The Square-Root Velocity Function (SRVF) method maps shapes in such a way that the shape space is a sphere. The distances among shapes can, then, be easily computed as great circles \cite{Joshi2007a,Kurtek2011,Srivastava16}  (Figure \ref{fig:methods}). Geometric Currents (GC) does something conceptually similar, transforming each shape into a mathematical function that can be represented as a point in the standard Euclidean vector space (Figure \ref{fig:methods}) based on Geometric Measure Theory \cite{Federer60}. Since this linear space is equipped with an Euclidean metric, it very easy to compute distances among shapes, and other standard statistical techniques such as PCA can also be used \cite{currents,Durrleman2009}. However, unlike SRVFs, it is not possible to transform the points in the new space back into the original shapes. Although the distances are now Euclidean, the GC transformation preserves much of the information present in the original shape space, whereas linear Geometric Morphometric methods simply ignore it \cite{currents}. Neither SRVF nor GC require Procrustes alignment.

Geometric Morphometric methods have been used in fields such as evolutionary biology \cite{Ferson1985,Bookstein}, medical image analysis \cite{Cootes94,Heimann09}, and archaeology \cite{Grosman2016,Macleod2018,Johaczi2018} for many years. Diffeomorphic methods, by contrast, have been applied only recently (e.g., \cite{dino_2011,bio_tut, RNA_2011, handwriting_2014,tree_2014,kimia_2015, animation_2017markus,tumour_2018,biomed_2020}). As far as we know these two, very different, approaches to shape analysis have not been tested against each other on real objects. Diffeomorphic methods should, in principle, provide better estimates of the true distances among real objects, but whether they do so in fact --- and whether any gain in accuracy justifies their greater computational cost --- is unclear. Given the range of shape variation among real objects, the assumption that shape space is linear may even be reasonable \cite{Klingenberg}. 

Here, then, we test one flavour of Geometric Morphometrics, semi-landmarks eigenshapes analysis, and three diffeomorphic methods --- LDDMM, SRVF and GC --- against each other in order to find out which of them performs best when classifying the shapes of real objects. The objects belong to three very different classes --- ancient Greek vases, the leaves of Swedish trees and gastropod shells --- chosen so that our results would be useful to archaeologists, botanists and zoologists, all of whom describe the shapes of the things that they study. 

Each of our datasets is divided into classes, for example, genera of shells.  Our test, then, rests on the ability of a statistical classifier, trained on distances computed by our various methods, to identify those classes. We show that, for all datasets, one diffeomorphic method --- SRVFs --- is superior to all other methods, including eigenshapes which, however, usually works impressively well. However, we also wanted to know what a good classification --- the kind that a trained human might make --- looks like, so we asked experts to undertake the same test. We find that most of our algorithmic methods beat the human experts. We conclude that such methods, particularly those that operate on curves rather than points, are very effective when applied to many shape classification problems, and can even be superior to humans. Finally, in homage to the grandfather of shape analysis, D'Arcy Wentworth Thompson, we show that some of these methods provide an answer to the problem that he posed in Chapter XVII of \emph{On Growth \& Form} \cite*{OGF}: how mathematics might be used to transform one shape into another.

\section*{Results}

\begin{figure}[ht]
\centering
\includegraphics[width=.8\linewidth]{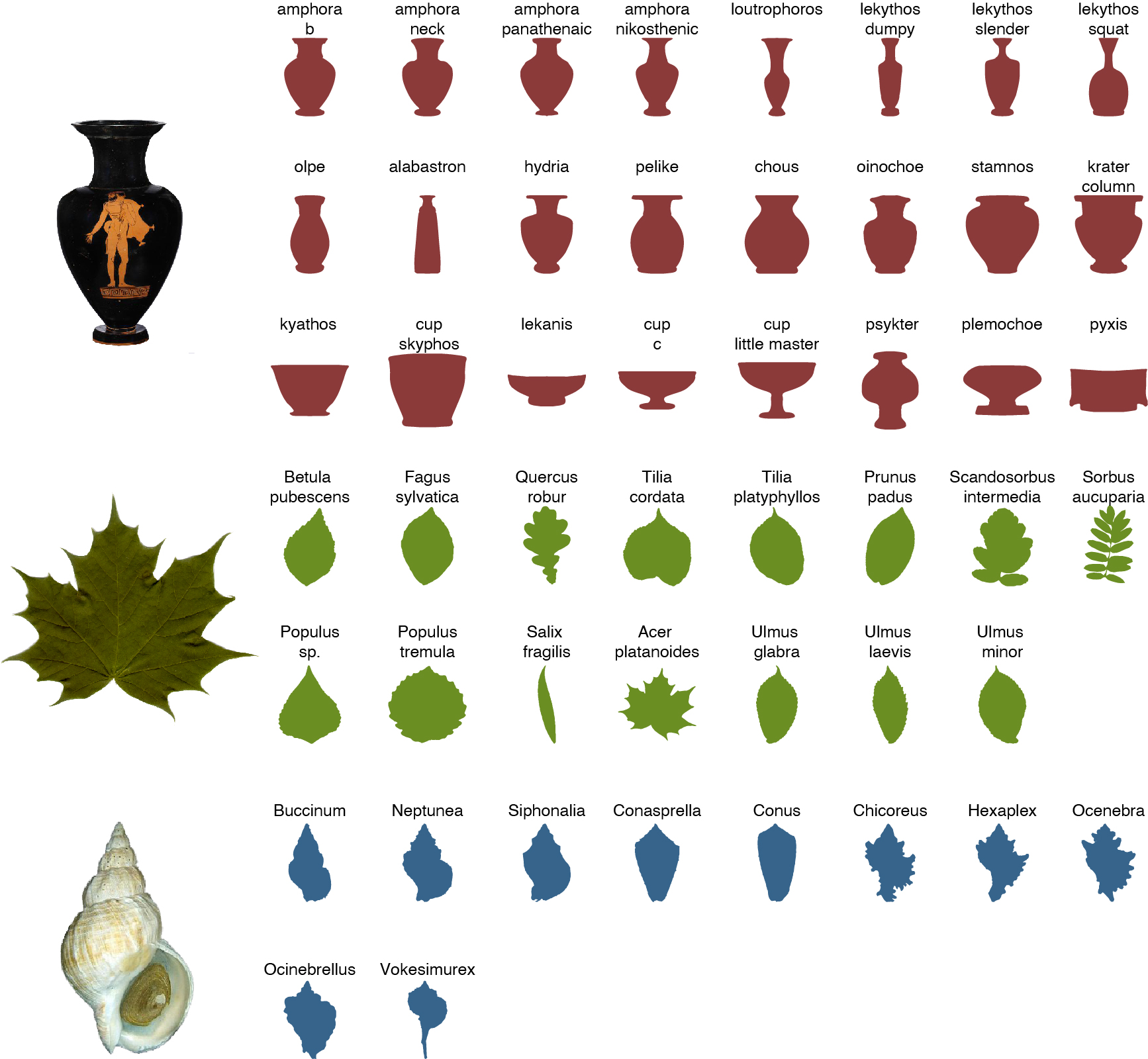}
\caption{Shape diversity in, from top to bottom, ancient Greek vases, Swedish leaves and Gastropod shells. \emph{Left}: An example of an original image for each class; \emph{right}: filled outlines of a randomly chosen member of each class within the three datasets.}
\label{fig:examples}
\end{figure}

We studied the two-dimensional outline shapes of three very different sets of objects: vases, leaves and shells. The vase outlines are based on 716 images of Athenian black- or red-figure vases classified into 24 classes: the shape categories used by vase scholars; the leaf outlines are based on 440 images of Swedish leaves classified into 15 Linnaean species; the shell outlines are based on 235 images of gastropod shells classified into 10 Linnaean genera. Figure \ref{fig:examples} shows, for each dataset, one of the original images from which outlines were extracted, as well as the outline of a randomly chosen member of each class. Our examples embrace a great variety of shapes. Where the outlines of Greek vases are mostly smooth, those of shells and leaves are often very jagged; and where our shells have quite similar aspect ratios, some leaves are needles, others are pancake-like, while others are something in-between. Within each class the individual objects are unique and distributed more-or-less evenly among classes.

\subsection*{Square-Root Velocity Functions are superior to other shape description methods}

To test our four shape description methods --- eigenshapes, LDDMM, SRVFs, and GC --- we first calculated the pairwise distances between all objects of each set --- vases, leaves and shells --- using each method.  We then trained a statistical classifier on the distances among a training set of between 51--67\% of the objects, and then asked the classifier to assign the remaining test objects to a class. To ensure that our results did not depend on the chance allocation of individuals among training and test sets, we constructed a hundred different sets by random stratified sampling and ran the classifier on each. Since the shape analysis methods compute distances between shapes, the obvious classifier is one that uses such distances directly, here the $k$-nearest neighbour ($k$-NN) classifier.  We measured classification success as the $F_1$-score, the harmonic mean of precision and recall of the obtained classification relative to ground-truth  \cite{MarslandBook2}.

\begin{figure}[ht]
\centering
\includegraphics[width=.8\linewidth]{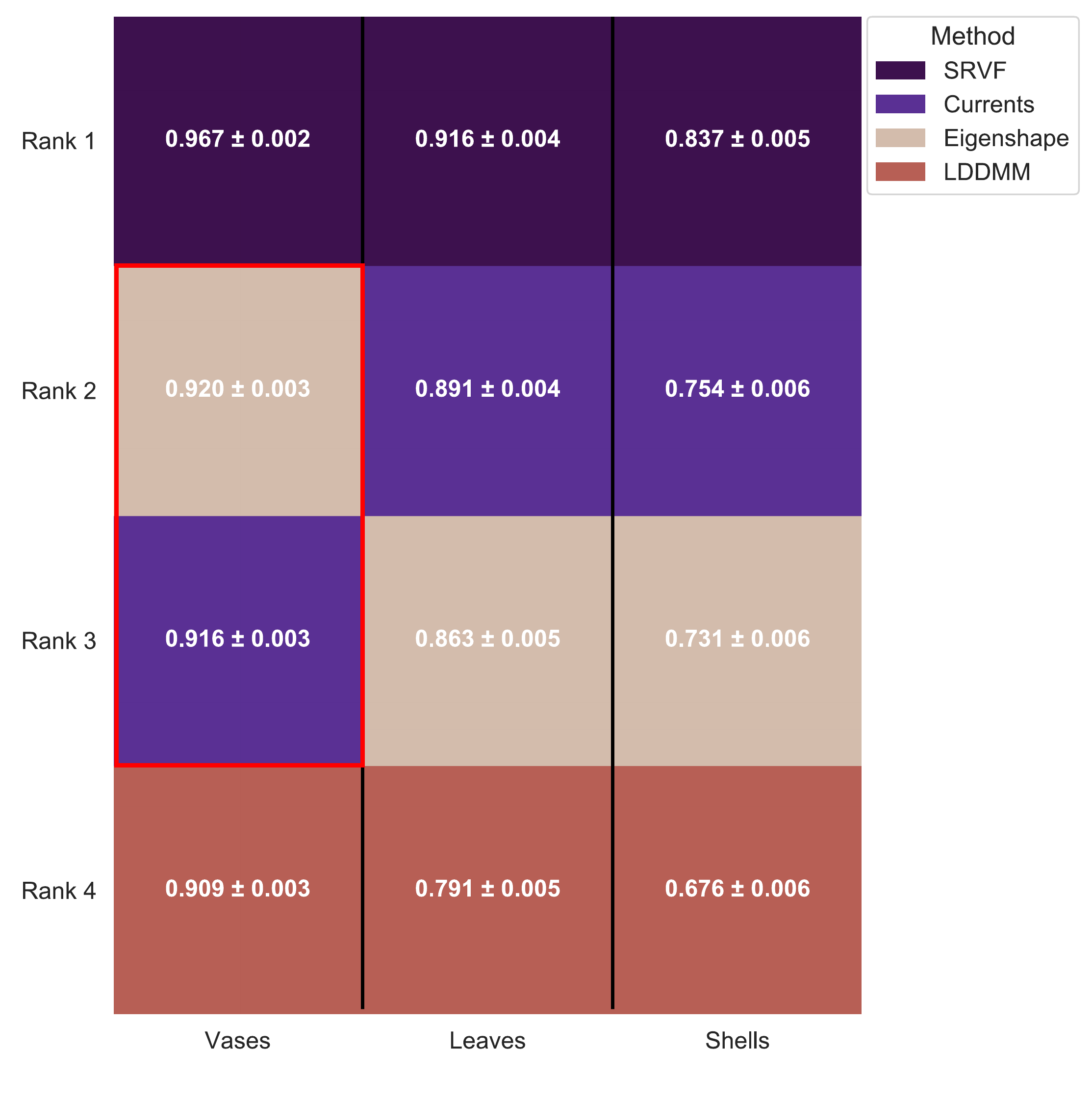}
\caption{Machine classification of three sets of objects using four shape analysis methods. Our measure of performance is  the $F_1$-score which captures the congruence between the $k$-NN's classification and ground truth.  The mean $F_1$-scores $\pm$ 95\% confidence intervals are based on randomly sampled training/test sets. The methods are ranked by decreasing $F_1$-scores within each object type. All methods showed statistically significant differences in performance from each other ($P< 10^{-6}$, two-tailed $t$-tests) except eigenshapes and Geometric Currents for vases as indicated by a red box ($P=0.1$). Each shape analysis methods has tuning parameters and, for each method and class of objects, we tried many combinations of them; only the results for the best-performing combination of parameter values on each object class are shown. The best method in all cases is SRVF, followed by Geometric Currents, but the linear method, eigenshapes, performs better than LDDMM.}
\label{fig:kNNresults}
\end{figure}

Even though the training sets were small --- a few hundred individuals divided among 10--24 classes --- the $k$-NN proved remarkably good at classifying outline shapes.  Its ability to do so, however, depended on the shape description method used. Figure \ref{fig:kNNresults} shows the ranked performance of each method over the object samples. The Square-Root Velocity Function method was the top-ranked method in all cases, being able to classify vases into their classes with 97\% accuracy, leaves with 92\% and shells with 84\% ($F_1$-scores); Geometric Currents performed next best overall, followed by eigenshapes. 
 
In order to show each method to its best advantage we varied their parameters (see Methods \& Materials); Figure \ref{fig:kNNresults} reports the best result for each class. The variation in performance that comes from tweaking parameters can be instructive. When trying eigenshapes, for example, we varied the number of principal components that went into the distances and found that, in all cases, the winner used at least 90\% of the total variance and, for vases 99.9\%, which suggests that some of the shape differences between classes are very subtle indeed.

\begin{figure}[ht]
\centering
\includegraphics[width=.8\linewidth]{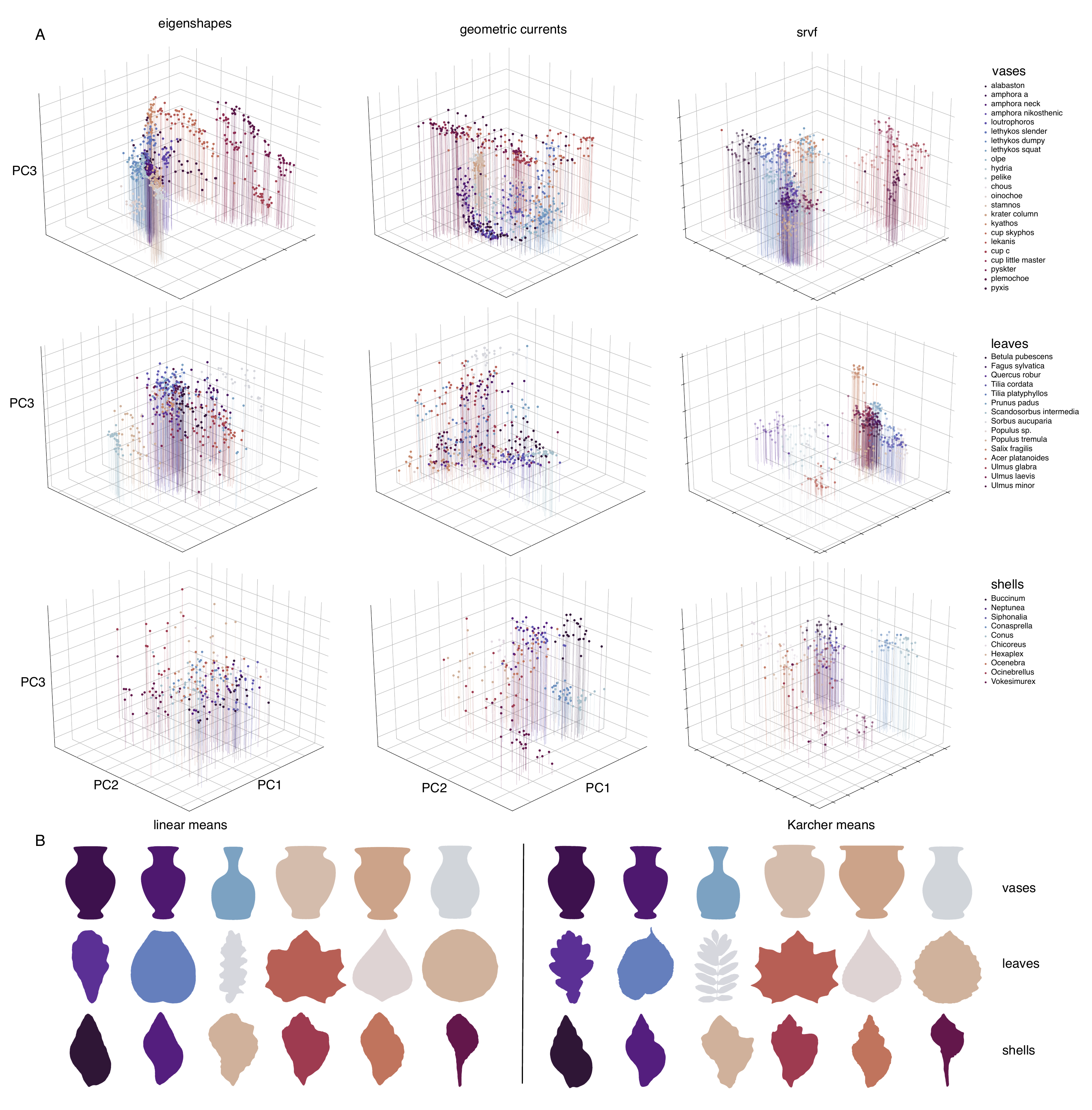}
\caption{\textbf{A.} Visualising the positions of objects in shape space. Objects plotted in the first three Principal Components of: \emph{left} eigenshapes (linear), \emph{centre} Geometric Currents, and \emph{right} SRVF space. Classes are colour-coded within each object type: \emph{top} vases, \emph{centre} leaves, \emph{bottom} shells. In general, the classes are increasing well separated moving from eigenshapes to Geometric Currents to SRVF space. \textbf{B.} Mean shapes for selected classes of objects. \emph{left}: linear means; \emph{right}: Karcher means. The Karcher means usually look more like the outlines of individual objects than the linear means do, the gain in detail being particularly pronounced for complex leaf and shell shapes.}
\label{fig:PCs}
\end{figure}

The best method, SRVFs, improves shape classification accuracy over the linear eigenshapes method by 5--10\% depending on the object class. However, the superiority of diffeomorphic methods is also evident when we plot the positions of the objects in the relevant shape space. Eigenshapes, Geometric Currents and SRVFs all yield principal components and, in general, the classes are better separated in GC and SRVF PC-space than they are in eigenshape space (Figure \ref{fig:PCs}A). 

We can also compute the average shape of each class using our various methods. For SRVFs this is the Karcher (Fr\'{e}chet) mean, an average shape estimated in Riemannian space \cite{Grove1973}. Figure \ref{fig:PCs}B shows that, where eigenshape means are rather amorphous, even blob-like, Karcher means retain more detail and so resemble the original objects much more closely (compare the objects in Figure \ref{fig:PCs}B with those in Figure \ref{fig:examples}). Thus our results show that a diffeomorphic shape description method, SRVFs, is better than the standard linear method, eigenshapes, at classifying the shapes of real objects and also at producing accurate averages of groups of objects. We note, however, that eigenshapes actually work surprisingly well and beat at least one diffeomorphic method, LDDMM, at least in our implementation. 

\subsection*{A machine shape-based classifier is superior to human experts}

We would like a machine classifier that classifies at least as well as humans do. But not all humans are equally adept at classifying all things. To find out how well our shape-based $k$-NNs perform we therefore formed a panel of experts composed of three classical vase scholars, three botanists, and three malacologists. We then asked each expert to classify a single set of objects into classes.  Each expert only classified objects about which they were an expert (i.e., the malacologists only got shells). Each expert was given outlines that resemble those in Figure \ref{fig:examples}; they had no direct information about the test set that the $k$-NN did not. All three experts were given the same set of object outlines to classify and told how many classes to make, but did not have to name them. Thus their task was the same as that given to the $k$-NN except that, instead of being trained on a training set, they had to rely on what they already knew. 

\begin{figure}[ht]
\centering
\includegraphics[width=.8\linewidth]{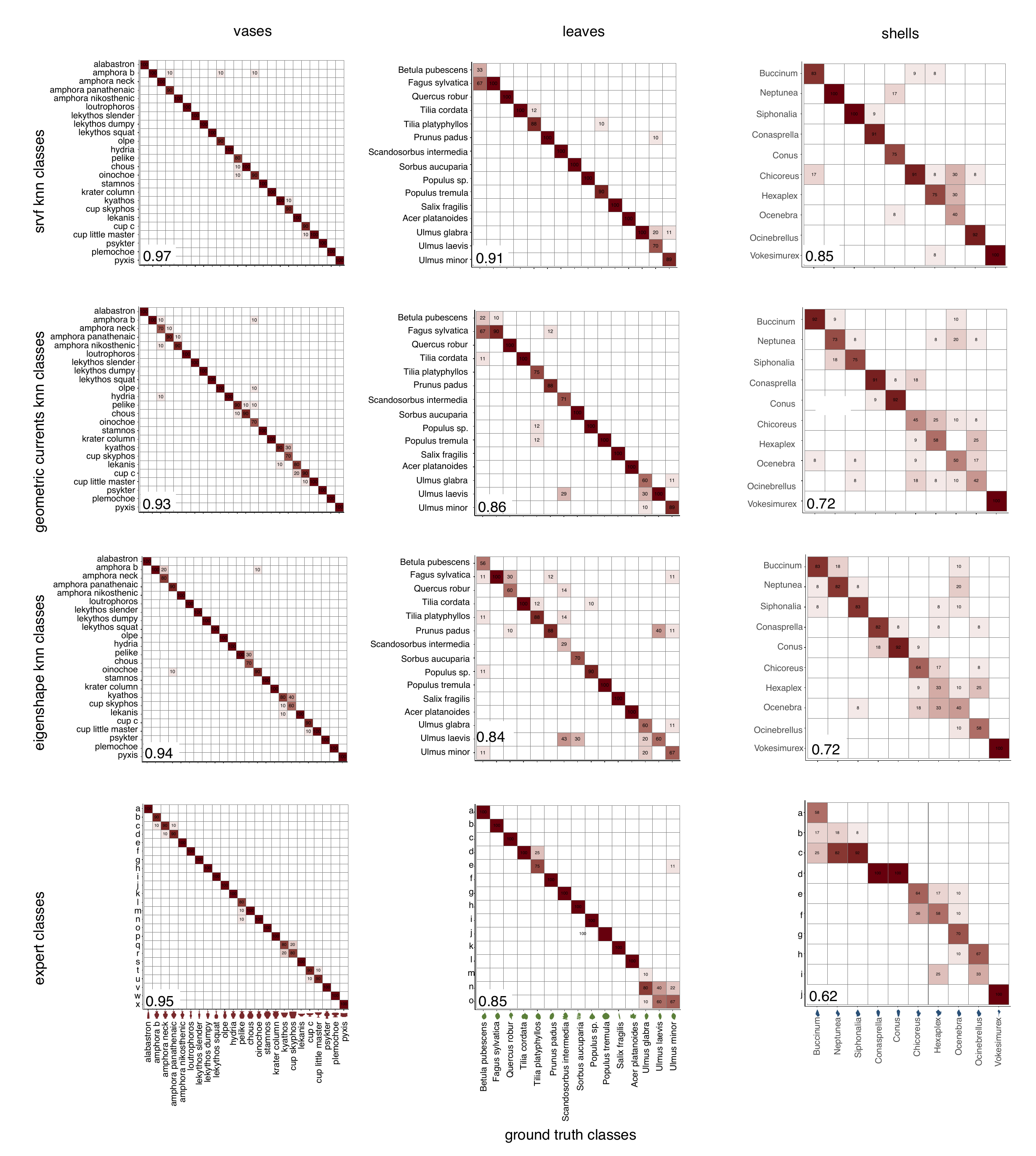}
\caption{Confusion matrices showing the ability of various methods to classify three classes of objects. Were the objects all perfectly classified, all elements would be placed on the diagonal, so off-diagonal elements are mis-classifications. The $F_1$ score of each classification is given in the bottom left corner of each panel. Top to bottom: classification by SRVF $k$-NN, Geometric Currents $k$-NN, eigenshapes $k$-NN, the best of the three experts for each dataset; left to right: classification of vases, leaves, shells. Note that the experts were asked to place outlines into similar shape groups, but not to label the groups.}
\label{fig:confusion}
\end{figure}

For these particular test sets, the SRVF-based $k$-NN classifier achieved $F_1$-scores of 0.971, 0.908, and 0.848,  for vases, leaves, and shells, respectively: comparable to the scores we found on our hundred-replicate data sets (Figure \ref{fig:kNNresults}). Our experts weren't as good: the mean $F_1$ scores of the three ($\pm$ one standard deviation) were 0.847 $\pm$0.087, 0.799 $\pm$0.039, and 0.574 $\pm$0.044 for the same objects. The best that any expert did on any dataset was 0.95 (for vases), but even that expert was beaten by the machine. Interestingly, the rank order of the average abilities of our expert groups --- vase-scholars $>$ botanists $>$ malacologists --- is the same as that of the machine classifiers, which suggests that the \emph{a priori} taxonomies of vases, leaves and shells that we used embody successively less shape information. Moreover, as the confusion matrices show, experts and algorithms tend to make the same kind of mistakes (Figure \ref{fig:confusion}). Where our experts tended to confuse kyathoi and skyphoi vases, the three species of \emph{Ulmus} leaves, and shells belonging to the muricid genera \emph{Hexaplex} and \emph{Chicoreus}, so did the algorithms. There are some differences. The SRVF-based $k$-NN correctly classified most \emph{Conus} and \emph{Conasprella} shells correctly even though they have very similar cone-shaped shells.  Our experts, by contrast, all failed to do so.  In general, however, our results suggest that, when classifying shapes, human experts and machine classifiers based on distances in shape space do much the same thing. It's just that algorithms do it better. 

\subsection*{Finding the shortest paths in shape space}

\begin{figure}
\centering
\includegraphics[width=.8\linewidth]{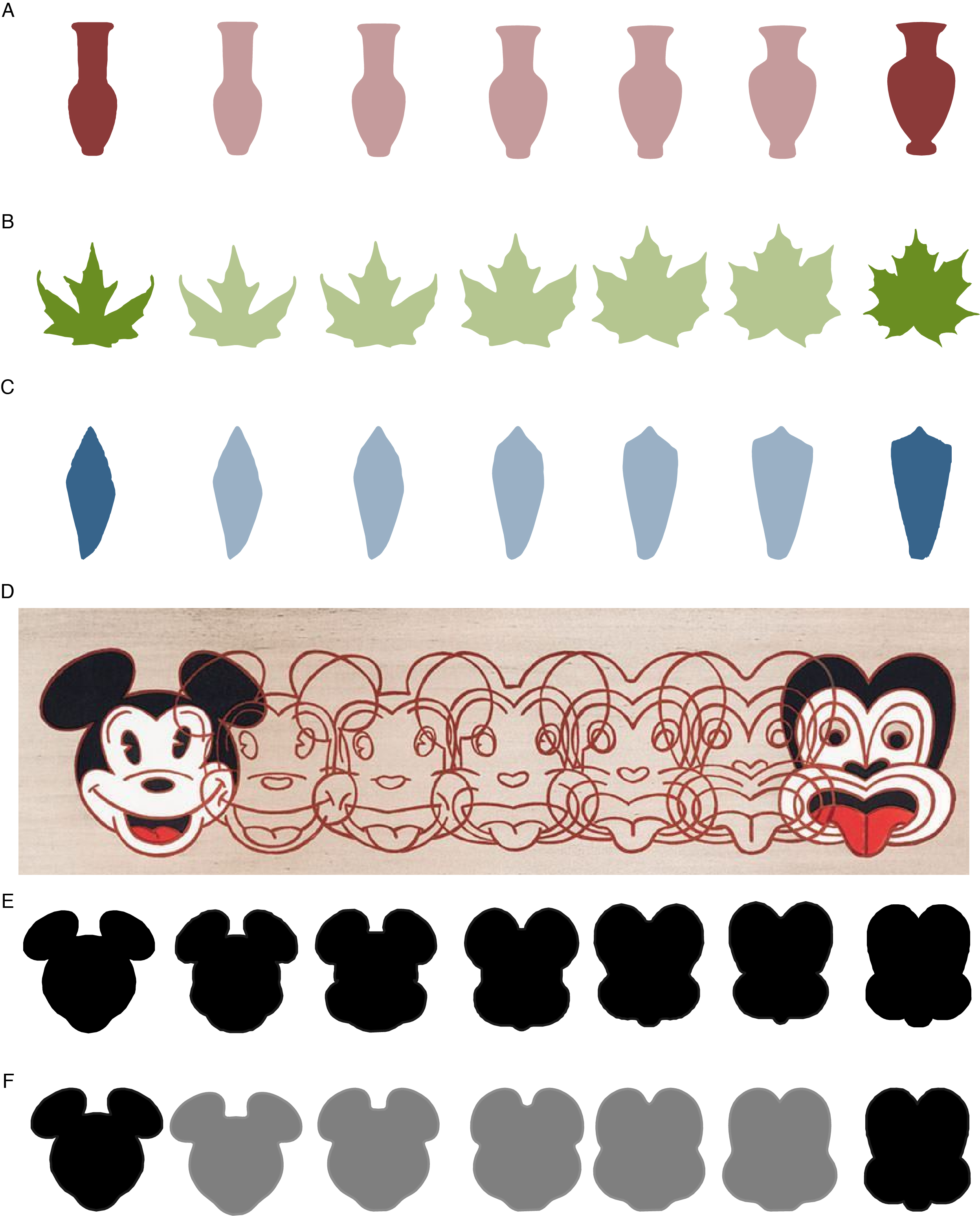}
\caption{Examples of shape transformations along geodesic paths. \textbf{A.} A Proto-Attic Neck Amphora (675-700 BCE) transformed, by five steps, into an Athenian Red Figure Neck Amphora (525-475 BCE); \textbf{B.} an early Miocene (20-18 Ma) maple leaf, \emph{Acer palaeosaccharinum} into the recent \emph{Acer platanoides}; \textbf{C.} The late Paleocene \emph{Hemiconus leroyi}  (59.2--56 Ma) into the recent \emph{Conus furvus}; \textbf{D.} \emph{Mickey to Tiki Tu Meke (1995)} by New Zealand artist Dick Frizzell; \textbf{E.} Frizzell's shapes isolated as outlines; \textbf{F.} Mickey into Tiki outlines as the shortest distance in SRVF shape space.}
\label{fig:transforms}
\end{figure}

SRVF and LDDMM work by transforming shapes into each other. When doing so, they find a geodesic --- the shortest path in shape space. Any point along this path can be back-transformed into a shape in the original space to produce a transformational series.  To illustrate this we transformed the outline of a plausible ancestor, or at least ancient relative, to one of our modern objects and inferred some intermediates.  Figures \ref{fig:transforms}A--C shows the transformation of a Proto-Attic Neck Amphora (725-675 BCE) into an Athenian Red  Figure Neck Amphora (525-475 BCE) \cite{Cook1997}; an early Miocene (20-18 Ma) maple, \emph{Acer palaeosaccharinum}  \cite{Denk2017} into the recent \emph{A. platanoides}; and the first known Conid gastropod, the late Paleocene \emph{Hemiconus leroyi}  (59.2--56 Ma) into the recent \emph{Conus furvus} \cite{Leroy2014, Tracey2017}. These examples are only illustrative: we do not claim that the earlier objects are true ancestors of the more recent ones. Indeed, the transformed objects need not be linked by evolutionary descent at all. In 1995 the New Zealand Pop artist, Dick Frizzell, transformed an American icon, Mickey Mouse, into a M\=aori one, the Tiki (Figures \ref{fig:transforms}D \& E). The SRVF geodesic path from Mickey to Tiki  is slightly different from the artist's  --- and 23\% more efficient (Figure \ref{fig:transforms}F). 
    
\section*{Discussion}

Although linear Geometric Morphometric methods have been widely used, they only approximate distances in the complex geometry of shape space \cite{Kendall1984, Klingenberg}. For biological objects this approximation largely suffices \cite{Klingenberg} and, consistent with this claim, we find that eigenshapes analysis generally performs quite well at our classification task. However, two diffeomorphic methods, SRVF and GC, are even better at distinguishing and classifying objects of different shape, and this is true for objects as disparate as vases, leaves and shells. In addition, the mean shape of groups of objects in these shape spaces also clearly preserve more detail than linear shape means do. These results imply that diffeomorphic methods, until now mostly studied by mathematicians, belong in the scientist's toolbox. All the implementations that we used are publicly available (see Material \& Methods). 

The three diffeomorphic methods do not perform equally well at the classification task. Since LDDMM does not simplify the shape space it might be expected that it would give the most accurate distances relative to human experts. In fact, of the three it performs the worst. This is because the metric used trades-off the precision of the transformation with the length of the path between them. For this reason it sometimes finds a transformation that is only close to the true target shape, and so may miss some of the finer distinctions among our classes. Since SRVF and GC work in spaces with much simpler geometries, they should be able to match curves exactly. However, implementation in a computer requires additional constraints. The GC algorithm first discretizes the shapes and, in doing so, sacrifices some information about them, while the SRVF algorithm avoids this at the cost of simplifying the path between the shapes. \emph{A priori} it is not clear which of these approaches would be most effective, but empirically SRVFs are for these datsets.

Our machine shape classifier worked best on vases, slightly less well on leaves and only moderately well on shells.  It may be supposed that these differences in performance depend on the shapes themselves, however, since the performance of the experts showed exactly the same rank order, it is much more likely that they depend on the quality of the classes. While our ground-truth classes were imposed by humans and chosen by us in the expectation that their members have, on average, different shapes, their natures vary. The vase classes are based on a scholarly taxonomy that largely depends on their gross shapes, but the leaf and shell classes are not; for modern biological genera and species are distinguished not only by gross shape gross shape, shape and positioning of constituent parts (e.g., spiral ribs, varices, leaf veins), but also microscopic, ecological, behavioural and genetic traits or abstract properties such as the ability to interbreed.  Even the differences among vase classes are not all visible from their outlines, depending, in part, on constructional details.  This means that the amount of information about class identity that is visible from shape outlines varies greatly among the three datasets.

Our classifiers used only shape rather than the many other features that might distinguish these groups.  Furthermore, the classifier that we used --- a $k$-NN --- requires very small training sets compared to the large training sets required by more sophisticated ML methods such as a Convolutional Neural Network (CNN). However, $k$-NN is the natural choice since our analysis gives distances among shapes rather than features.  Even so, the success of our shape-based classification is remarkable. We imagine that they might be useful for the automatic classification of the innumerable objects that differ in shape, not only those we have studied here, but even things as diverse as protein structures, the spectrograms of bird songs or the melodies of pop songs (e.g, \cite{Urbano2011, Imai2016, Srivastava2016, Cope2012}).  Given suitable data our methods could also be applied to three-dimensional shapes (e.g., \cite{Koutsoudis2011,Johaczi2018}). That would be useful for rotationally asymmetrical objects, but also require much more computational effort.
 
Our classifier was more accurate than the judgements of experts, almost regardless of the shape-analysis method under the hood.  Why is this?  We asked our experts and found that they were often led astray by prior knowledge. Where the machine classifier was trained to distinguish the groups actually present, the experts sometimes sought the groups that they thought \emph{should} have been there.  For example, all three malacologists failed to distinguish between the closely related genera \emph{Conus} and \emph{Conasprella}. They did so because the classification of the Conidae remains unsettled \cite{Puillandre15}, and the relationship between shell shape and genera unclear.  Indeed, one of our experts had second thoughts about the cones, gave us a revised classification before being told the ground-truth, and got the best $F_1$-score among the malacologists, 0.716. Our intention, however, is not to diminish experts who, after all, usually have much more information about the objects that they classify, but rather show how effective machine shape-classifiers can be, even when based on very small training sets.

Our study revealed some limitations of the diffeomorphic methods as currently implemented.  The first is that, compared to linear methods, they are computationally expensive. In the implementations we used, a SRVF or LDDMM registration for a single pair of shape outlines takes, on average, 1--3 seconds to process on a modern laptop. Computing all 255,970 pairwise distances for our Greek vase data set of 716 objects takes, then, 85 hours if performed sequentially. The Geometric Currents algorithm is much faster and takes about 10 seconds to complete the same task, although it does suffer from some memory issues. However, eigenshapes --- Procrustes alignment, PCA, and distance calculations --- takes, on average, only 1.5 seconds.

A second limitation is more profound. The curves that we have used are closed --- without a start or end --- and the algorithms can rotate one shape relative to the other to find lower energy paths from one to the other. In general, this should result in homologous parts being aligned to each other, but it need not. Instead, the spire of one shell might be aligned to another's siphonal canal or the neck of one vase to another's base. Indeed, when using SRVFs to transform various shapes into each other we came across some instances of just this phenomenon. That may not matter for the purposes of mere classification, but any evolutionary interpretation of the distances would be incorrect, for the inferred path would be one that evolution could not possibly have taken. Geometric morphometric methods, which depend on the correspondence of explicitly homologous points --- landmarks --- are not vulnerable to this error. Constraining the curve transformations by adding some landmarks may solve this problem, and we will consider this in the future.

More than a hundred years ago D'Arcy Wentworth Thompson posited his ``theory of transformations'' which held that species closely related by evolutionary descent should also be related by ``simple'' shape transformations; and that ``small'' transformations indicate particularly close evolutionary affinities \cite{MandM, OGF}. To demonstrate this Thompson relied on outline drawings of an animal, adding a rectilinear grid that was deformed using a regular transformation, with the image of the animal deformed along with it until it more closely matched another animal. Our modern equivalents dispense with the grid and match the curves more accurately. However, in spirit they are the same, and our transformations illustrate how the evolution of shape in Riemannian space can be modelled so that it might be mapped onto a phylogeny or even used to infer one \cite{Gavryushkina2014, Parins2017}.

%%%%%%%%% Stuff taken out of the discussion

%But eigenshapes is extremely fast: assembling the matrix of covariances of points for all 716 shapes and finding its eigenspectrum takes less than a second. 

%The use of the distances directly in the $k$-NN classifier has the benefit that the classifier can be trained using relatively few training examplars for each class. This is particularly noticeable when one considers methods such as Convolutional Neural Networks and other gradient-descent-based machine learning algorithms. However, those methods can be applied to the original images, while for shape methods it is necessary to extract the curve outline first. While this has been an area of research interest for a long time in computer vision --- and is something that humans do easily --- there do not yet exist completely reliable methods \cite{De16}.

.

%All of the computational methods we use produce symmetric distances, at least approximately: the distance from A to B is the same as the distance from B to A. While this seems like a reasonable assumption, there is evidence that humans do not do this. For example 

%Our semi-landmark method is not complete, in the sense that we do not report a form with optimisation of landmark locations. We did try this, and it made no difference to the results despite considerable computational cost. 

%However, the semi-landmarks do have one benefit in practice: the first and last points are clearly delineated, which is not true for the curves. The curve-matching methods have the freedom to rotate the curve arbitrarily, which could mean that the lip of one vase is aligned with the base of another. While this might produce a shorter distance than a more standard match, this flexibility may not be helpful in practice, since the homologous points are informative. 

\matmethods{\subsection*{Datasets}
The vase images were obtained from the \href{https://www.beazley.ox.ac.uk/index.htm}{Beazley Archive Pottery Database} (BAPD) at Oxford University; their taxonomy, which was modified slightly from the standard shape taxonomy given in the BAPD, was checked by two experts, TM and DRP.
The leaf images are based on the \href{https://www.cvl.isy.liu.se/en/research/datasets/swedish-leaf/}{Swedish Leaf Dataset} previously used in the image analysis and shape literature \cite{swedish_leaf}; the images came with species labels which were checked by an expert, TER. The shell images were obtained from \href{http://gastropods.com}{Gastropods.com}; the images came with species labels whose taxonomy was standardized to the \href{http://www.marinespecies.org}{World Register of Marine Species (WoRMs)} and checked by AML. Each image represents a unique object and was checked to ensure that it was complete and in standard orientation. The sources of the original images are given in a datafile on this repository [url].

\subsection*{Data Preparation}

Shape methods require an outline of the object, and often it is necessary to extract this from a digital photograph. While this has been an area of research interest for a long time in computer vision --- and is something that humans do easily --- there do not yet exist completely reliable methods \cite{De16}. We used a common contour extraction algorithm, the Marching Squares method \cite{maple} on a binarized version of each image, with the threshold chosen experimentally. For the leaves and shells no other pre-processing was performed, but for the vases the handles were removed using a spline fit, which was verified and, if necessary, corrected manually. The vases were also made to have a reflective symmetry through a central vertical axis by computing the outline contour of each side, and using the shorter one of the pair, reflecting it to make the full shape. This removes structures such as the spouts of pouring vessels.

Each outline curve was sampled to have an identical number of equally spaced points --- 139 for the vases, 150 for the shells, and 200 for the leaves --- by sampling a  cubic spline fitted to the curve. Preliminary experiments showed that, at these resolutions no difference between the interpolated curve and the original shape were visible to the naked eye. The point sets were aligned using Procrustes alignment to remove the global transformations of scale, rotation, and translation from the curves. This is necessary for the linear method and LDDMM, but not for SRVF or GC. Examples of the resulting shape outlines with filled interiors are shown in Figure \ref{fig:examples}. The same datasets of shape outlines were used when testing all methods. 
The shape outline data are available at the following DOIs: Vases: \url{10.6084/m9.figshare.14551002}, Leaves: \url{10.6084/m9.figshare.14551005}, Shells: \url{10.6084/m9.figshare.14551044}.
.

\subsection*{Estimating distances}

Parameters for each method were chosen experimentally based on the training data, and the upper-triangular distance matrix between all pairs of shapes computed for each method. \emph{Eigenshapes}: We used the points that parameterise the curve as semi-landmarks. We experimented with optimising the position of these landmarks, but it was computationally expensive and did not improve the results. We computed the principal components of the point coordinates of all shapes and, from these, the Euclidean distances among them using the first $d$-dimensions, where $d$ was chosen based on the amount of the variance explained, ranging from $0.75$ to $0.999$.  \emph{LDDMM}: We used the implementation described in \cite{langevin_2017} available \href{https://github.com/tonyshardlow/reg_sde}{\textit{here}}, running for 20 timesteps. \emph{SRVF}: We used the implementation available \href{https://github.com/jdtuck/fdasrsf_python}{\textit{here}}. The  Path-Straightening algorithm is described in \cite{path_st_2011} and available \href{https://fdasrsf-python.readthedocs.io/en/latest/}{\emph{here}}. The algorithm transforms one shape to another in $\kappa \geq 2$ steps. The output is the geodesic distance, which is the inner product in SRVF space between the first shape and the final shape in the transformation. To compute our distance matrix, we set $\kappa=2$. \emph{Geometric Currents}: we used the method described by \cite{currents} available \href{https://github.com/olivierverdier/femshape}{\textit{here}}. This implementation takes three parameters: a non-negative integer, $s$, determining the size of the matrix representation; the mesh-size, $m$; and a scaling parameter, $\sigma \geq 0$. We tested three options for each parameter where $ 1 \leq s,\sigma \leq 4$ and $16 \leq m \leq 24$.

\subsection*{Machine classification}

Most machine learning algorithms take as input features of the elements of the dataset (or their complete representation), rather than distances. We, however, used our various shape analysis methods to compute distances among objects and wish to classify on those. For this reason we implemented our own $k$-nearest neighbour ($k$-NN) classifier that takes a distance matrix as its input.  Our $k$-NN assigns elements of the test set to the class of the majority of the closest $k$ points in the training set, where $k$ is a user-selected parameter. We tested values of $k$ between 3 and 12 for each method and object class and found the $k$ that results in the highest $F_1$-score. We ran the $k$-NN  on 100 randomly selected samples from the training sets of each dataset and computed the $F_1$-scores, where the samples were selected with a pseudo-random number generator. For vases the ratio of training:test set was 480:236, leaves 300:140, and shells 120:115. In order to ensure that training set size was not the reason for better performance in the case of vases, we also reduced the size of the training set in that case (to 10 in each class), leaving the test set alone, without significantly changing the results. Interestingly, even when reducing the training set further, to 2 in each class, the classifier still did well. We used the \href{https://scikit-learn.org/stable/modules/generated/sklearn.metrics.f1_score.html}{\texttt{sklearn}} implementation of the $F_1$-score with the \textit{average} parameter set to ``weighted''. 

\subsection*{Expert classification}

Each expert was given a standard test set of shape outlines as individual images and asked to partition them into $n$ groups, where $n$ is the number of ground-truth classes, by sorting them into folders. The objects were anonymized so that no expert had any information about them that the machine classifier did not. The experts were not asked to identify the groups that they formed. Each expert's classification was then compared to the ground-truth classification by an $F_1$-score. 

\subsection*{Transformations}

To create the transformation plots seen in Figure \ref{fig:transforms}, we used the SRVF Path-Straightening algorithm with $\kappa = 5$. Note that the transformations are not necessarily symmetric even if the shapes themselves are symmetric, such as Mickey and Tiki. Therefore, to display a symmetric transformation between Mickey and Tiki, we split the outlines in half and transformed these halves from one to another. The transformations were then reflected and attached. Furthermore, to test the efficiency of our transformation with the artist's, albeit in a metaphorical sense, we computed the sum of the distances between consecutive outlines, i.e., the energy needed to deform one shape into the other.

}

\showmatmethods{} % Display the Materials and Methods section

\acknow{We thank Thomas Denk, Swedish Museum of Natural History, for images of fossil leaves; Steven Tracey, Natural History Museum, London, for images of fossil cones; Keith Kirby, Plant Sciences, University of Oxford, for help with the plant classification task, and Dick Frizell for the use of his art work \emph{Mickey to Tiki Tu Meke}. AS-J was supported by an  EPSRC studentship, awarded to Brunel University London.}

\showacknow{} % Display the acknowledgments section

% Bibliography
\bibliography{shaperefs}

\end{document}